\documentclass{article}

\usepackage{epsfig}
\usepackage{subfigure}
\usepackage{calc}
\usepackage{amssymb}
\usepackage{amstext}
\usepackage{amsmath}
\usepackage{amsthm}
\usepackage{multicol}
\usepackage{natbib}
\usepackage{graphicx}

\newcommand{\field}[1]{\mathbb{#1}}
\newcommand{\ALGO}{MUCCA}

\begin{document}

\title{A Scalable Multiclass Algorithm for Node Classification}
\author{Giovanni Zappella}
\maketitle

\begin{abstract}

We introduce a scalable algorithm, \ALGO\, for multiclass node classification in weighted graphs. Unlike previously proposed methods for the same task, \ALGO\ works in time linear in the number of nodes. Our approach is based on a game-theoretic formulation of the problem in which the test labels are expressed as a Nash Equilibrium of a certain game. However, in order to achieve scalability, we find the equilibrium on a spanning tree of the original graph. Experiments on real-world data reveal that \ALGO\ is much faster than its competitors while achieving a similar predictive performance. 
\end{abstract}

\section{Introduction}
Classification of networked data is a quite attractive field with applications in computer vision, bioinformatics, spam detection and text categorization. In recent years networked data have become widespread due to the increasing importance of social networks and other web-related applications. This growing interest is pushing researchers to find scalable algorithms for important practical applications of these problems. 
\\
In this paper we focus our attention on a task called \textit{node classification}, often studied in the semi-supervised setting~\citet{labprop}. Recently, different teams studied the problem from a theoretic point of view with interesting results. For example \citet{treeopt,WTA, shazoo} developed on-line fast predictors for weighted and unweighted graphs and Herbster et al. developed different versions of the Perceptron algorithm to classify the nodes of a graph (\citet{HerbPerc,HerbPercOld}). 

\citet{Pelillo} introduced a game-theoretic framework for node classification. We adopt the same approach and, in particular, we obtain a scalable algorithm by finding a Nash Equilibrium on a special instance of their game. The main difference between our algorithm and theirs is the high scalability achieved by our approach. This is really important in practice, since it makes possible to use our algorithm on large scale problems.

\section{Basic Framework}\label{basicframework}

Given a weighted graph $G=(V,E,W)$, a labeling of $G$ is an assignment $y=(y_1,..,y_n) \in \{0,1,...,c\}^n$ where $n=|V|$.\\
We expect our graph to respect a notion of regularity where adjacent nodes often have the same label: this notion of regularity is called \textit{homophily}.
Most machine learning algorithms for node classification (\citet{HerbPerc,WTA,labprop, shazoo}) adopt this bias and exploit it to improve their performances.\\
The learner is given the graph $G$, but just a subset of $y$, that we call training set. The learner's goal is to predict the remaining labels minimizing the number of mistakes.

\citet{WTA} introduce also an irregularity measure of the graph $G$, for the labeling $y$, defined as the ratio between the sum of the weights of the edges between nodes with different labels and the sum of all the weights.
Intuitively, we can view the weight of an edge as a similarity measure between two nodes, we expect highly similar nodes to have the same label and edges between nodes with different labels being ``light''.
Based on this intuition, we may assign labels to non-training nodes so to minimize some function of the induced weighted cut.

In the binary classification case, algorithms based on min-cut have been proposed in the past (for example \citet{mincutblum}). Generalizing this approach to the multiclass case, naturally takes us to the \textit{multi-way cut} (or multi-terminal cut --- see \citet{surveycut}) problem. Given a graph and a list of terminal nodes, find a set of edges such that, once removed, each terminal belongs to a different component. The goal is to minimize the sum of the weights of the removed edges.\\
Unfortunately, the multi-way cut problem is MAX~SNP-hard when the number of terminals is bigger than two (\citet{18survey}). Furthermore, efficient algorithms to find the multi-way cut on special instances of the problem are known, but, for example, it is not clear if it is possible to reduce a node classification problem on a tree to a multi-way cut on a tree.

\section{Graph Transduction Game}\label{game}

In this section we describe the game introduced by \citet{Pelillo} that, in a certain sense, aims at distributing over the nodes the cost of approximating the multi-way cut. This is done by expressing the labels assignment as a Nash Equilibrium. We have to keep in mind that, since this game is non-cooperative, each player maximizes its own payoff disregarding what it can do to maximize the sum of utilities of all the players (the so-called social welfare).
The value of the multi-way cut is strongly related to the value of the social welfare of the game, but in the general case a Nash Equilibrium does not give any guarantee about the collective result.\\
%
%
In the Graph Transduction Game (later called GTG), the graph topology is known in advance and we consider each node as a player. Each possible label of the nodes is a pure strategy of the players.
Since we are working in a batch setting, we will have a train/test split that induces two different kind of players:
\begin{itemize}
\item \textbf{Determined players}($I_D$) those are nodes with a known label (train set), so in our game they will be players with a fixed strategy (they do not change their strategy since we can not change the labels given as training set)
\item \textbf{Undetermined players}($I_U$) those that do not have a fixed strategy and can choose whatever strategy they prefer (we have to predict their labels)
\end{itemize}

The game is defined as $\Gamma=(I,S,\pi)$, where $I=\{1,2,...,n\}$ is the set of players, $S=\times_{i \in I} S_i$ is the joint strategy space (the Cartesian product of all strategy sets $S_i \subseteq \{1,2,...c\}$), and $\pi: S \to \field{R}^n$ is the combined payoff function which assigns a real valued payoff $\pi_i(s) \in \field{R}$ to each pure strategy profile $s \in S$ and player $i \in I$.

A mixed strategy of player $i \in I$ is a probability distribution $x$ over the set of the pure strategies of $i$. Each pure strategy $k$ corresponds to a mixed strategy where all the strategies but the $k$-th one have probability equals to zero.

We define the utility function of the player $i$ as
$$u_i(s) = \sum_{s \in S} x(s) \pi_i(s)$$
where $x(s)$ is the probability of $s$.

We assume the payoff associated to each player is additively separable (this will be clear in the following lines). This makes GTG a member of a subclass of the multi-player games called polymatrix games.
For a pure strategy profile $s=(s_1,s_2,...s_n) \in S$, the payoff function of every player $i \in I$ is:
$$\pi_i(s)= \sum_{j \sim i} w_{ij} \field{I}_{\{s_i = s_j\}}$$
where $i \sim j$ means that $i$ and $j$ are neighbors, this can be written in matrix form as
$$\pi_i(s) = \sum_{j \sim i} A_{ij} (s_i,s_j)$$
where $A_{ij} \in \field{R}^{c \times c}$ is the partial payoff matrix between $i$ and $j$, defined as $A_{ij} = I_c \times w_{ij}$, where $I_c$ is the identity matrix of size $c$ and $A_{ij} (x,y)$ represent the element of $A_{ij}$ at row $x$ and column $y$.
The utility function of each player $i \in I_U$ can be re-written as follows:
\begin{center}
\begin{tabular}{ r l }
$ u_i(x) $	&$= \sum_{i \sim j} x_i^T A_{ij} x_j$ 	\\
	 	&$= \sum_{i \sim j} w_{ij} x_i^T x_j$	\\
	 	&$= \sum_{i \sim j} w_{ij} \sum_{k=1}^c x_{i_k} x_{j_k}$
\end{tabular}	
\end{center}	
where $k$ is an action selected from the player's set and in case $i$ is a determined node with training label $k$, $x$'s components will be always zeros except the $k$-th corresponding to the pure strategy $k$.
Since the utility function of each player is linear, it is easy to see that players can achieve their maximum payoff using pure strategies.

In a non-cooperative game, a vector of strategies $S_{NE}$ is said to be a (pure strategies) Nash Equilibrium, if $\forall i \in I$, $\forall s_i' \in S_i : s_i' \neq s_i \in S_{NE}$, we have that 
$$ u_i (s_i, S_{NE}^{-i}) \geq u_i (s_i', S_{NE}^{-i})$$
where $u_i (s_i, S^{-i})$ is the strategy configuration $S$ except the $i$-th one, that is replaced by $s_i$. In practice, no player $i$ will change its strategy $s_i$ to an alternative strategy improving its payoff.
\\
There are no guarantees that the Nash Equilibrium exists in pure strategies, but \textit{ any game with a finite set of players and finite set of strategies has a Nash Equilibrium in mixed strategies} (\citet{nasheq}, also see \citet{agtbook}). 
In this case each player does not have to choose a strategy but it mixes its choices over its strategies. Instead of maximizing its payoff, it will maximize its expected payoff.

Abusing of terminology, in the following sections we may talk about labels or pure strategies with the same meaning.

\subsection{The Evolutionary Stable Strategies approach}

\citet{Pelillo} propose to find a Nash equilibrium of the GTG using the Evolutionary Stable Strategies. We briefly present their approach in order for the reader to better understand the difference between their algorithm and ours.

The evolutionary stable strategies (ESS) approach is well known (\citet{evolution}) in the game-theoretic literature.
It considers a game played repeatedly; each repetition of the game is seen as a generation, where an imaginary population evolve through a selection mechanism that, at each step, gives to the best ``choices'' a growing portion of the total population. 

The algorithm (later called GTG-ESS), at each generation, updates the probability associated to every action $h$ of every player $i$ as
$$ x_{ih} (t+1) = x_{ih}(t) \frac{u_i(e_h)}{u_i(x(t))} $$
The previous formula is just the discrete version of the so-called multi-population replicator dynamic:
$$ \dot{x}_{ih} = x_{ih} (u_i(e_h, x_{-i}) - u_i(x))$$
where $e_h$ is a vector of zeros except the $h$-th component that is one and $x_{ih}$ is the $h$-th strategy of player $i$.
The fixed points of the previous equations are Nash Equilibria, and the discrete version has the same properties --- for further details see \citet{Pelillo}.

In this case, the computational cost of finding the Nash Equilibrium is $O(k |V|^2)$ where $k$ is the number of iterations and considering the number of classes as a constant factor.
\citet{Pelillo} experimentally found that the number of iterations grows linearly with the number of nodes, so they consider the running time close to $O(|V|^3)$, but they do not seem to have any upper bound on the number of iterations.

\section{The $\ALGO$ algorithm}

Use GTG-ESS for large scale networks can not be considered a viable solution, even if the time complexity of the algorithm were to be demonstrated in the order of $\Theta (|V|^3)$.
A possible alternative is to apply some known results about regret minimization, such as those described by \citet{cesa}, in order to converge to a weaker notion of equilibrium, for example the Correlated Equilibrium (\citet{correlated}). Unfortunately, the results of our experiments with the Correlated Equilibrium were not satisfactory.

In this section we present $\ALGO$: a multiclass Classification Algorithm. Our algorithm consists in finding a Nash Equilibrium of the Graph Transduction Game on a special graph: a tree. We will show that in this way we achieve both good accuracy and scalability.
In the remaining part of this section we will assume that the graph $G$ is a tree. 

Now, we briefly introduce few notions those will be useful later in this section (see Figure~\ref{fig:1}):

\begin{figure}
\centering
\includegraphics[width=70mm]{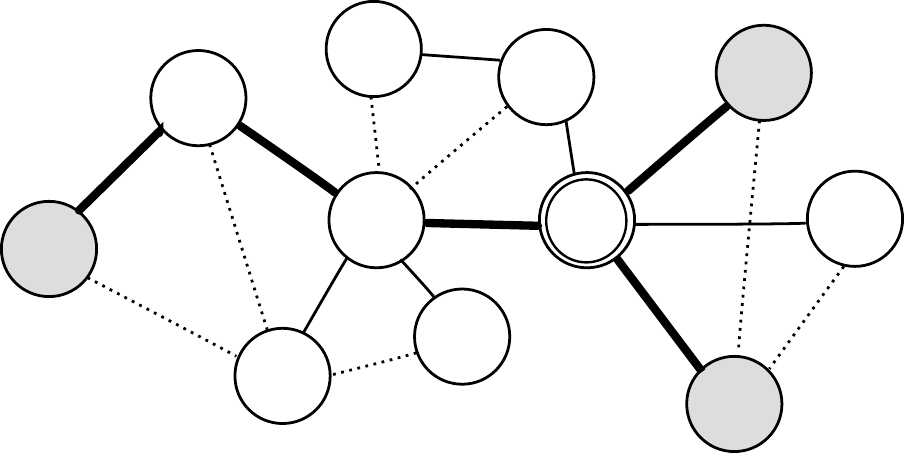}
\caption{Some of the elements introduced in this paragraph: grey nodes are labeled nodes, the node with two circles is a fork, the fat black edges are those flagged ``black line'', the thin solid black line edges are edges of the spanning tree and the dotted edges are edges of the original graph not selected for the spanning tree.}\label{fig:1}
\end{figure}

\begin{itemize}
\item \textbf{Revealed node}: a node whose label is known
\item \textbf{Unrevealed node}: a node whose label is unknown
\item \textbf{Fork}: an unrevealed node that is connected to at least 3 revealed nodes by edge-disjoint paths
\item \textbf{Hinge node}: a revealed node or a fork
\item \textbf{Hinge tree}: component of the forest created by removing from $G$ all the edges incident to hinge nodes
\item \textbf{Native hinge tree}: component of the forest created by removing from $G$ all the edges incident to revealed nodes. Its connection nodes are intended to be only labeled nodes.
\item \textbf{Hinge line}: a path connecting two hinge nodes such that there are no internal nodes those are hinge nodes
\item \textbf{Connection Node}: an hinge node that is adjacent to a node in an hinge tree
\item \textbf{$\epsilon$-edge}: given $\mathcal{P}_{ij}$, the path between $i$ and $j$, an $\epsilon$-edge is $\epsilon_{ij} \in \arg\min_{e \in \mathcal{P}_{ij}} w_e$, where $w_e$ is the weight of the edge $e$. 
\item \textbf{Grafted tree}: a tree without hinge nodes that is connected to just one node on an hinge line
\end{itemize}

$\ALGO$ works in four phases:

\begin{enumerate}
\item Mark all the paths between revealed nodes and find all the fork nodes
\item Estimate the label of each fork node
\item Assign a label to all the nodes on the hinge lines using a min cut technique
\item Assign a label to all the remaining nodes
\end{enumerate}

\begin{figure}
\begin{center}
\includegraphics[width=60mm]{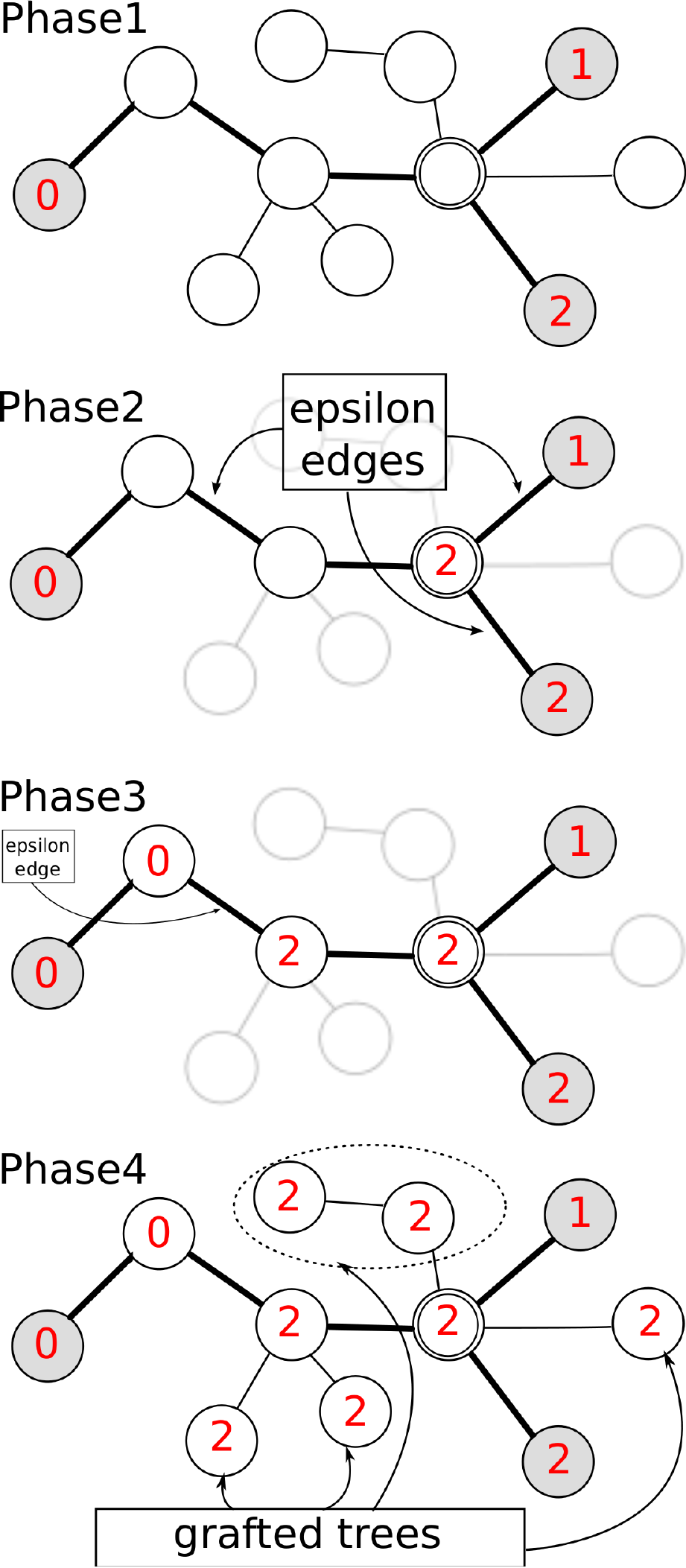}
\end{center}
\end{figure}

Now we describe what it does in every phase:
\\
\textbf{Phase 1}\\
Starting from each revealed node, it does a breadth-first search until another revealed node or a leaf are found. Then if a revealed node was found, during the backtracking phase, $\ALGO$ marks the edges on the path to the starting node with a special flag that we will call ``black line''. 
After that, each node with more than 3 edges marked as ``black line'' incident to it, is a fork.

\textbf{Phase 2}\\
Given a native hinge tree $\mathcal{H}$ that contains the fork $\mathcal{F}$, we can categorize its connection nodes in $c$ categories using their labels. For each path between $\mathcal{F}$ and each connection node of $\mathcal{H}$, we have an $\epsilon$-edge as defined before. The label assigned to $\mathcal{F}$ is the same as the category (of the connection nodes) that has the maximum sum of weights over the distinct $\epsilon$-edges on the path between $\mathcal{F}$ and the nodes of that category.

\textbf{Phase 3}\\
On every hinge line, we label the nodes using min cut: in case the hinge nodes at the beginning and at the end of the line has the same label, all the nodes on the hinge line will be labeled with that label. Otherwise, all the nodes before the $\epsilon$-edge are assigned with the label of the node at the beginning of the line, and the others with the label of the node at the end of the line.
In case we have more than one edge with the same weight of the $\epsilon$-edge (for example all the edges have the same weight), we use nearest neighbor to choose the labeling on the line.

\textbf{Phase 4}\\
All the remaining nodes are in grafted trees. In this case, we assign the label of the node on the hinge line (connected to the tree) to all the nodes in that particular grafted tree.\\
\\
We now prove that $\ALGO$ finds a Nash Equilibrium for this special case of the GTG.\\
\\
\textbf{Theorem.} \textit{The labeling found by $\ALGO$ is a Nash Equilibrium of the Graph Transduction Game when the graph is an undirected tree.}
\\
\textbf{Proof.}
As we explained in Section 3, a profile of strategies $S_{NE}$ is a Nash Equilibrium if no one has incentive to deviate from its strategy.
This means that $\forall i \in I $,  $ u_i(s_i, S_{NE}^{-i}) \geq u_i(s_i', S_{NE}^{-i})$.
For the purpose of contradiction suppose that exists a node $j$ such that it can improve its payoff by changing its strategy.
\\
$j$ can not be contained in a grafted tree (those labeled in phase 4) since all the nodes contained in those trees have the same labels, so, whatever is the label, each of them gets a payoff of $\sum_{i \sim j} w_{ij}$, the maximum possible payoff.
\\
$j$ can not be on an hinge line since they are labeled using min cut, so in the best case the payoff of each node is already the maximum payoff; in the worst case the payoff is the maximum minus the weight of the $\epsilon$-edge. Since the $\epsilon$-edge, by definition, has the minimum weight, there are no chance to improve the payoff.
\\
$j$ can not be a revealed node (for obvious reasons).
\\
$j$ can not be a fork. Since we use min cut to label the hinge lines if the $\epsilon$-edge of an hinge line is not incident to the fork, the node adjacent to the fork on that hinge line will have the same label of the fork. In this way the fork will get the part of payoff given by the edge between it and the adjacent node. Even if the $\epsilon$-edges are incident to the fork, the label prediction can not achieve a payoff better than the one achieved by the majority vote.
\\
Since $j$ can not be a revealed node, nor a fork, nor a node on an hinge line, nor a node on a tree with just a connection node, it can not be in $G$.\hfill$\square$
\\
\\
Since we consider the number of forks as a constant factor, $\ALGO$ runs in $O(|V|)$. Note that the labels of the unlabeled nodes of every native hinge tree can be predicted just using information about that singular native hinge tree. In this way, once the tree is splitted in native hinge trees, the predictions for the labels contained in every subtree are independent from the other subtrees. So, they can be computed using different threads, processes or even machines and we just need to get back a list of points (node id, predicted label).

\section{Experiments}\label{results}

In this section we present our experimental results. Since binary classification is a (well-studied) special case of our problem, we first compare $\ALGO$ with the state of the art in this particular field. In the second part of this section we test our algorithm versus its competitors on multiclass problems.
We have to keep in mind that $\ALGO$ works in linear time and can be used on large-scale graphs, where we can not test its competitors.

Our \textbf{experimental protocol} is quite simple: for every size of the training set (and possibly on every class), we did 10 runs for each algorithm. Algorithms working on trees were ran on: Minimum Spanning Trees since previous experimental works showed that predictors get their best results on this kind of tree (\citet{shazoo}), and on Random Spanning Trees (RandomSpanningTrees) generated as in \citet{Wilson} to test the most scalable solution possible.
We tested $\ALGO$ also in a committee version: in the tables numbers before the predictor's name represent the number of predictors in the committee (for example $n$*MyPred will represents a committee of $n$ predictors MyPred).
Each predictor in the committee predicts its own labels using its own tree and then we aggregate the predictions with a majority vote over the committee.

In order to better understand the competitiveness of $\ALGO$, we will compare it with some well-know \textbf{algorithms} besides GTG-ESS.

\textbf{Label Propagation} (abbreviated LABPROP), introduced by \citet{labprop}, is a popular algorithm for node classification and one of the most accurate algorithms in the literature.
LABPROP computes a real-valued function $f: V \to \field{R} $ on $G$, and then assigns a labeling to $G$ using the values of $f$. The algorithm minimizes the following quadratic energy function: 
$$E(f) = \frac{1}{2} \sum_{i \sim j} w_{ij} (f(i)-f(j))^2$$
with constrains on the labeled nodes.
Its running time is $O(|E| \times |V|)$.

\textbf{SHAZOO} is a quite new algorithm introduced by \citet{shazoo} and, at the best of our knowledge, it is the most accurate scalable algorithms between our competitors. SHAZOO has strong theoretical foundations, but the idea behind the algorithm is really different from our one: it is an online algorithm based on nearest neighbor. 
\begin{figure}[h]
\centering
\includegraphics[width=70mm]{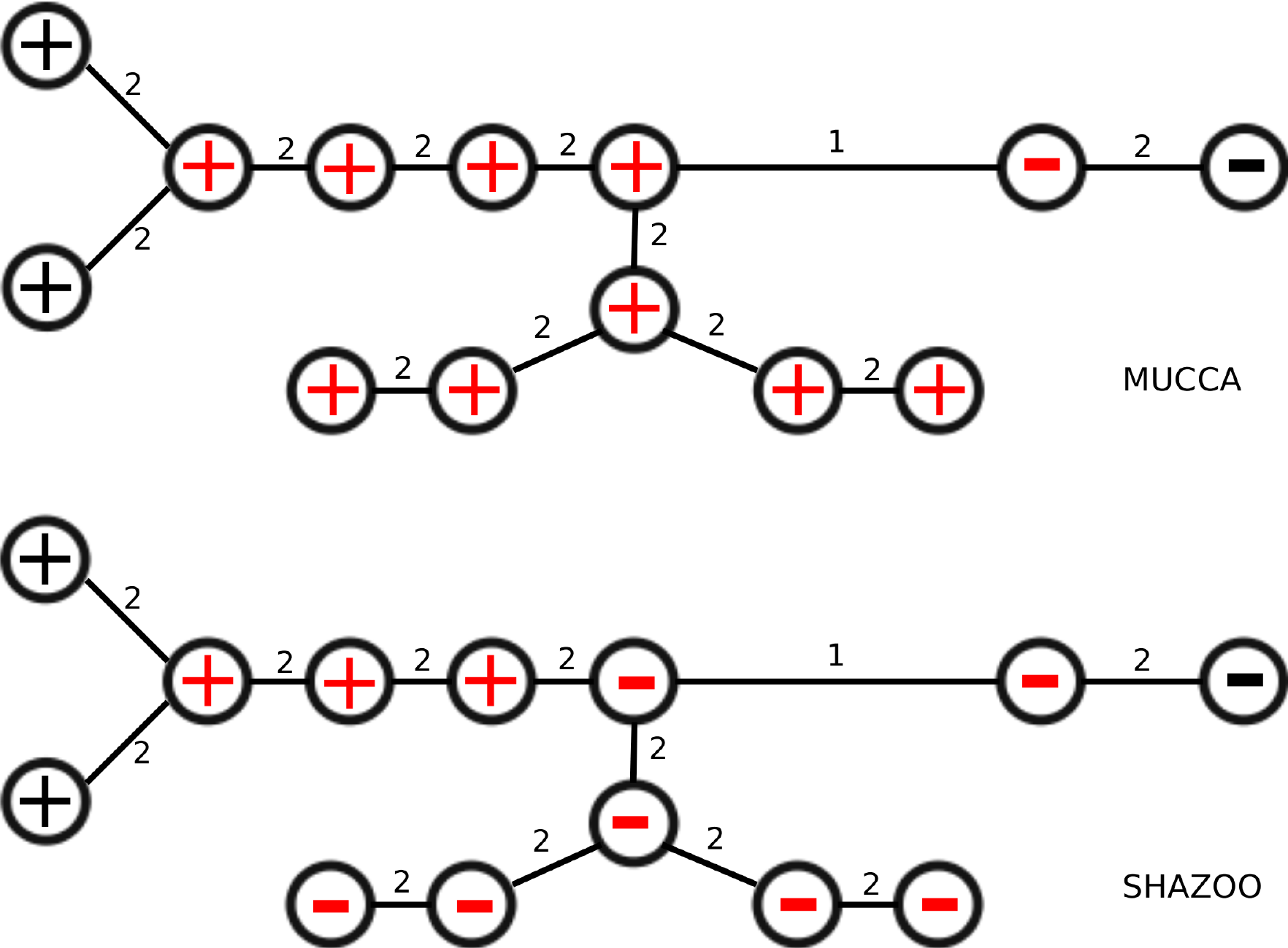}
\caption{Red labels are predicted respectively by $\ALGO$ in the first tree and SHAZOO in the second one}\label{fig:2}
\end{figure}
In Figure 2, you can see a toy example where SHAZOO and $\ALGO$ behave in a different way: $\ALGO$ predicts all the nodes on the left hand side as positive since there are two nodes labeled as positive, connected to a fork that is labeled as positive and the $\epsilon$-edge is the one with weight 1. Clearly every player wants to maximize its payoff, so the algorithm produces a labeling with a minimum possible cut. In the second part, for SHAZOO the node in the middle looks closer to the negative example than to the positive ones. In this way also all the nodes contained into the grafted tree will be labeled in a different way.\\
SHAZOO's running time is $O(|V|)$.

\textbf{Weighted Majority Vote} (later called WMV) predicts the label of a node $i$ using a majority vote on the labeled neighbors weighted on the edges connecting them to $i$. Its running time is $\Theta(|E|)$.

Both LABPROP and GTG-ESS can be used for multiclass datasets but it is not clear if it is possible to modify SHAZOO in order to get a multiclass algorithm. In the multiclass section we replace SHAZOO with WMV. 
We did not include in our comparison WTA (\citet{WTA}) since it was always outperformed by SHAZOO. Graph Perceptron (\citet{HerbPerc}) was omitted since it requires a lot of computational resource and performed poorly in previous comparisons (for example in \citet{WTA}).

\subsection{Binary classification}

In order to create a fair comparison, for the experimental activity we will generate our datasets in the same way of \citet{shazoo}.

\subsubsection{Datasets}

We choose two real-world and well-known datasets to generate our graphs: USPS and RCV1.
\\
\textbf{USPS} is a set of hand written digits collected by the United States Postal Service, it contains 9298 images of $16 \times 16$ pixels (gray scale); the dataset called \textbf{RCV1} is a subset of 10000 articles in chronological order taken from Reuters Corpus (a huge collection of news released by \citet{}). All the articles have been pre-processed using TF-IDF.
Both datasets were natively multiclass, so we tested our binary predictors using a standard one-vs-rest schema. We have 10 binary experiments for USPS and 4 binary experiments for RCV1.

We generated our graphs with as many nodes as the total number of examples for each dataset, keeping for each node only 10 nearest neighbor (before symmetrization) using the Euclidean distance $\| x_i - x_j \|$. Edges' weights have been calculated as
$$ w_{ij} = e^{-\| x_i - x_j \| / \sigma^2 }$$
where $\sigma$ is the average between the weights of all the edges incident to $i$ or $j$.

\subsubsection{Experimental results}

    \begin{table*}[h]

    \centering
     \tabcolsep=0.1cm
     
\scriptsize
\begin{tabular}{l|c|c|c|c|c|c|c|c}
&\multicolumn{4}{c|}{ RCV1 }&\multicolumn{4}{c}{ USPS }\\
&0.5\%&1\%&2\%&5\%&0.5\%&1\%&2\%&5\%\\
\hline
$\ALGO$+MinimumSpanningTree&27.70&25.34&23.12&18.85&3.24&2.06&1.35&1.10\\
$\ALGO$+RandomSpanningTree&32.02&29.41&27.06&22.83&9.20&7.49&5.78&4.26\\
\hline
SHAZOO+MinimumSpanningTree&26.87&24.35&21.64&18.26&2.92&1.85&1.29&1.08\\
SHAZOO+RandomSpanningTree&30.77&28.16&25.70&21.99&8.72&7.26&5.49&4.17\\
\hline
3*$\ALGO$+RandomSpanningTree&28.30&25.25&22.95&18.75&6.65&5.29&3.80&2.46\\
3*SHAZOO+RandomSpanningTree&27.29&24.44&22.24&18.44&6.67&5.52&3.84&2.48\\
\hline
7*$\ALGO$+RandomSpanningTree&26.09&23.17&20.62&16.61&6.19&4.60&3.14&1.78\\
7*SHAZOO+RandomSpanningTree&25.71&22.80&20.55&16.61&6.50&4.97&3.32&1.86\\
\hline
11*$\ALGO$+RandomSpanningTree&25.39&22.40&19.92&15.97&6.12&4.43&2.90&1.59\\
11*SHAZOO+RandomSpanningTree&25.30&22.37&20.01&16.09&6.46&4.85&3.15&1.69\\
\hline
\hline
GTG-ESS&21.66&19.13&17.00&14.37&2.69&1.67&1.17&0.92\\
LABPROP&26.74&23.51&20.84&16.37&7.34&5.27&3.65&2.36\end{tabular}

    \caption{RCV1 and USPS - Error rate of each algorithm. Results are averaged over the 4 and 10 binary problems.}
    \label{t:binario}
    \end{table*}


The results, shown in Table~\ref{t:binario}, are not conclusive, but we can observe some interesting trends:
\begin{itemize}
\item GTG-ESS is the most accurate algorithm, but its computational complexity makes it not particularly suitable for most of the current practical applications.
\item $\ALGO$'s accuracy is good and it is always close or better that SHAZOO's one.
\item LABPROP approaches the competitors as the training set size grows.
\end{itemize}

\subsection{Multiclass classification}

In the multiclass comparison we replaced RCV1 with two other datasets, since it is not just multiclass but also multi-label (an element has more than one label). 
As we explained before, SHAZOO works just on binary problems, so the only scalable competitor in this section is WMV.

\subsubsection{Datasets}
\begin{itemize}
\item USPS is the same graph we used for the binary classification.
\item CARDIO \textit{consists of measurements of fetal heart rate (FHR) and uterine contraction (UC) features on cardiotocograms classified by expert obstetricians}. We generated a graph starting from the feature vectors, in the same way we did for USPS and RCV1. The graph is constituted by 2126 nodes and 13696 edges. The nodes are divided in 3 classes (fetal states)\footnote{The Cardiotocography dataset was created by \citet{fetal} and now is freely available on the UCI website with a complete description: http://archive.ics.uci.edu/ml/datasets/Cardiotocography}.
\item GHGRAPH is a graph created from the data released by Github.com for their contest in 2009. Every node represents a repository and an edge means that there is a developer working on both repositories connected by that edge. In case more than one developer works on those repositories, the number of developers is used as a weight for that edge. The label of each repository is the most used programming language to write the code in that repository. We kept only the biggest connected component of the graph that includes 99907 nodes (64885 of them are labeled) and 11044757 edges. Each programming language is a different class and at the end we had 40 classes.
\end{itemize}

\subsubsection{Experimental results}

    \begin{table*}[h]
     \tabcolsep=0.1cm
\tiny
\begin{tabular}{l|c|c|c|c|c|c|c|c|c|c|c|c}

&\multicolumn{4}{c|}{ USPS }&\multicolumn{4}{c|}{ CARDIO }&\multicolumn{4}{c}{ GHGRAPH }\\
&0.5\%&1\%&2\%&5\%&0.5\%&1\%&2\%&5\%&0.5\%&1\%&2\%&5\%\\
\hline
$\ALGO$+MinimumSpanningTree&19.14&8.5&6.84&5.49&31.32&28.96&26.94&27.58&-&-&-&-\\
$\ALGO$+RandomSpanningTree&48.92&39.84&30.54&20.6&34.88&32.54&30.01&29.66&65.70&62.01&57.68&51.37\\
\hline
7*$\ALGO$+RandomSpanningTree&31.79&18.67&12.23&7.87&34.44&28.03&25.13&24.32&55.33&51.14&47.56&43.55\\
11*$\ALGO$+RandomSpanningTree&29.56&14.89&10.06&6.8&33.99&26.82&24.86&23.98&51.88&48.64&46.25&42.92\\
\hline
WMV&84.28&78.96&69.28&47.48&64.47&62.35&57.05&47.08&74.03&67.50&60.15&50.37\\
\hline
\hline
LABPROP&43.73&16.22&9.01&5.82&28.24&24.42&23.32&22.21&-&-&-&-\\
GTG-ESS&14.37&7.75&5.28&4.6&40.65&36.91&28.81&26.29&-&-&-&-\\
\end{tabular}
  \caption{Multiclass experiments - Error rate of each algorithm in function of the train size. Please note that we did not run GTG-ESS and LABPROP on GHGRAPH due to limited computational resources.}
    \label{t:multi}
\end{table*}



The results of our experiments, shown in Table~\ref{t:multi}, are not conclusive, but we can observe some interesting trends:
\begin{itemize}
\item It is not really clear which one between GTG-ESS and LABPROP is the most accurate algorithm, but anyway $\ALGO$ is always competitive with them.
\item $\ALGO$ is always much better than WMV. As expected WMV works better on ``not too sparse'' graphs such GHGRAPH, but even in this case it is outperformed by $\ALGO$.
\item GTG-ESS and LABPROP's time complexity did not permit us to run them in a reasonable amount of time with our computational resources.
\end{itemize}

\section{Conclusions and future work}\label{conclusions}
We introduced a novel scalable algorithm for multiclass node classification in arbitrary weighted graphs.
Our algorithm is motivated within a game theoretic framework, where test labels are expressed as the Nash equilibrium of a certain game. In practice, \ALGO\ works well even on binary problems against competitors like Label Propagation and SHAZOO that have been specifically designed for the binary setting.

Several questions remain open. For example, committees of \ALGO\ predictors work well but we do not know whether there are better ways to aggregate their predictions. Also, given their common game-theoretic background, it would be interesting to explore possible connections between committees of \ALGO\ predictors and GTG-ESS.

\bibliographystyle{plainnat}
{\small
\bibliography{simple}}

\end{document}